\documentclass{article}
\usepackage{ijcai22}

\usepackage{times}
\usepackage{soul}
\usepackage{url}
\usepackage[hidelinks]{hyperref}
\usepackage[utf8]{inputenc}
\usepackage[small]{caption}
\usepackage{graphicx}
\usepackage{amsmath}
\usepackage{amsthm}
\usepackage{booktabs}
\usepackage{algorithm}
\usepackage{algorithmic}
\usepackage{dirtytalk}
\usepackage{xcolor}

\newtheorem{theorem}{Theorem}

\title{Efficient Customer Service Combining Human Operators and Virtual Agents}
\author{ Yaniv Oshrat, %\textsuperscript{\rm 1}
    Yonatan Aumann, %\textsuperscript{\rm 1}
    Tal Hollander,
    Oleg Maksimov, %\textsuperscript{\rm 1}
    Anita Ostroumov, %\textsuperscript{\rm 1}
    Natali Shechtman, %\textsuperscript{\rm 1}
    Sarit Kraus \\%\textsuperscript{\rm 1} \\
    Department of Computer Science, Bar-Ilan University\\
    }

\begin{document}

\maketitle

\begin{abstract}

The prospect of combining human operators and virtual agents (bots) into an effective hybrid system that provides proper customer service to clients is promising yet challenging. The hybrid system decreases the customers' frustration when bots are unable to provide appropriate service and increases their satisfaction when they prefer to interact with human operators. Furthermore, we show that it is possible to decrease the cost and efforts of building and maintaining such virtual agents by enabling the virtual agent to incrementally learn from the human operators. We employ queuing theory to identify the key parameters that govern the behavior and efficiency of such hybrid systems, and determine the main parameters that should be optimized in order to improve the service. We formally prove, and demonstrate in extensive simulations and in a user study, that with the proper choice of parameters, such hybrid systems are able to increase the number of served clients while simultaneously decreasing their expected waiting time and increasing satisfaction.
\end{abstract}

\section{Introduction}
Virtual agents (bots) are increasingly used for providing on-line assistance to customers, even for complex or moderately complex tasks. The deployment of these virtual agents raises two main challenges. First, customers frequently would like to interact with humans \cite{Clutch,Forbes}, especially when they are frustrated that the virtual agent is not capable of providing an appropriate service \cite{ashfaq2020chatbot}. The second difficulty is the cost and domain knowledge required when initially deploying the agent and the constantly required updating during its lifetime \cite{BlurPrint,Enterprisers}. To address these two challenges, we propose a holistic approach of virtual agents and human operators working together to provide satisfactory service. 
The goal is that the virtual agents will incrementally learn from human operators and be able to handle the majority of the requests quickly and efficiently, while the human operators are used for those requests that cannot be attended this way and for cases in which clients specifically seek human assistance. Implementing this configuration can provide a speedy, inexpensive service, attending more clients with fewer human operators, and providing satisfactory service without extensive need for explicit domain knowledge acquisition and integration. 

In this paper we study the design of such hybrid customer-service systems, both theoretically and experimentally, establishing that, with proper design and choice of parameters, such systems can indeed offer significant performance improvements. We present an efficient methodology to develop such systems, and, in particular, the virtual agent, in a manner that requires little domain expertise. We show that such an agent, though relatively easy to develop, can significantly improve the service (measured by both  customer waiting times and the number of human operators required to operate the service), while maintaining customer satisfaction.

The paper is built as follows: Section \ref{sec:Problem_Definition} defines the problem. In Section \ref{sec:our_model} we present the model, define its various parameters, explore their trade-offs and determine ways to optimize them. Section \ref{sec:Model_Analysis} presents theoretical analysis of the model. We then demonstrate how to construct and design such systems (Section \ref{sec:Designing and Integrating the Virtual Agents: Methodology and Human Evaluation}): Employing an experimental service environment that we implemented, we provide a methodology for developing a virtual agent using off-the-shelf ML tools, which requires only limited explicit human programming. We present experiments that we conducted with human subjects, testing their actual performance and their results. Finally, we present results of extensive simulations that we conducted to test the system, with multiple customers, virtual agents and human operators, in various configurations (Section \ref{sec:Simulation}).

Our main contribution is to exhibit the efficiency and utility of such hybrid customer service systems. In particular, we demonstrate, theoretically and experimentally, that with the proper choice of parameters it is possible to significantly increase the number of customers the system can handle while also decreasing the expected waiting time of the clients. Additionally, we present efficient and simple methods for the design of the necessary virtual agents, and demonstrate that they can handle 70\% of the questions just by learning from human experts. This, in turn, allows the number of attended clients to significantly increase without increasing the clients' waiting time. 
%A believe more advanced agent can achieve a higher increase. 

\section{Related Work}

\paragraph{Motivation of Hybrid Systems for customer service}
Virtual agents for maintaining client service (\say{chat bots}) are very common today, and may be found in various businesses, from pizza stores to health care services \cite{koole2021practice}. Naturally, there are numerous tools and software packages that assist in building chat bots 
\cite{IBM,Maruti,Viraj}. 
When conversational bots emerged, it looked like an appealing idea to replace all human operators by bots, and to provide customer service by bots only \cite{nath2018chat,rizk2020unified}. Nevertheless, the experience of recent years, with many such services, shows that the bot-only service is not mature yet \cite{ashktorab2019resilient}, and that people tend to prefer service that enables them to receive human assistance when it is needed, even though parts of the service are automated \cite{paikens2020human,ashfaq2020chatbot}. This understanding led to focus on various service schemes that combine conversational bots with human operators, and to study the optimal ways to perform an efficient hand-off from bot to human \cite{rozga2018human}.  

It should be noted that there are other configurations, which are based on providing service to clients by human operators only, but augment the operators with virtual tools such as an advising agent \cite{aviv2021advising}, intent detection and entity extraction capabilities \cite{paikens2020human}. However, in this paper we refer to schemes in which the virtual agent communicates directly with clients, and in certain conditions it transfers the control to human operators. 

We claim that in the situations that we consider there is no need to conduct a long conversation with the customer in order to answer relatively easy questions. Furthermore, from previous questions answered by the human operator, our agent learns to incrementally improve its ability to answer question. The main challenge is the small number of examples and the cold start problem. Other attempts to determine such similarities given a relatively small number of questions were described, for example, in \cite{mccreery2020effective,ma2019domain}). Another approach is to use general trained models \cite{minaee2021deep,sun2020analysis} such as Sentence-BERT \cite{reimers-2019-sentence-bert}.
Our formal model can be used to determine the benefit of introducing a more sophisticated virtual agent that can converse with the customer and possibly answer complex questions.

\paragraph{Modeling customer service centers}
There have been many attempts made to formally model quality in service \cite{sankar2020various} and call centers \cite{koole2002queueing}, and to develop call center simulations \cite{avramidis2005modeling}
in order to provide staffing and scheduling of human operators \cite{tezcan2014routing,legros2019scheduling}. After the introduction of chat communication, the modeling focuses on the ability of a single human operator to attend more than one client simultaneously, in contrast to voice channels in which this parallel approach is usually impossible \cite{enqvist2017chat,long2018customer}. We focus on chat communication and models centers that are staffed with both virtual agents and human operators. We present both formal analysis and simulation results that are based on human experimental data.

\section{{Problem Definition and Goals}}
\label{sec:Problem_Definition}
We consider a virtual service center which provides help to clients who are engaged in performing a complex activity that consists of a list of tasks. The tasks are sequential and may depend on each other, i.e. in order to be able to perform a certain task there may be a need to successfully perform a previous task(s). For example, when setting a Wi-Fi connection, one needs to assemble the parts of the routing gear (router, cables, adapters) and to define user data (user name, password) with the ISP. Only after having these components figured individually can one assemble them together to get a working home Wi-Fi connection. 

When clients encounter difficulties in performing the tasks, they may contact the service center for help. Human operators attend help requests according to their order of arrival. An operator has to understand the situation (i.e. what the specific task is, what causes the problem) and to provide instructions to the client in order to solve the problem. 

%Sometimes the problem is obvious from the current situation (Q: 'This cable won't fit in the hole.', A: 'You took the wrong type of cable, you should take the other one.'), but there are situations in which the current problem is a result of a mistake or a malfunction performed by the client in previous tasks (A: 'You plugged in the short cable instead of the long one, therefore it can't reach the PC.'). The operator has to find the source of the current problem and guide the client to the solution.

There are many examples of such situations in real-life. For example, ready-to-assemble furniture vendors %, like IKEA or Walmart, who
support their clients while they try to %correctly  
assemble products they bought, and cell phone vendors %who need to 
guide clients in the process of inserting a flash memory card into their cell phone to make it work. 
%Home Internet service providers (ISPs) that need to guide new clients in the process of establishing a working connection to the internet (unboxing gear, connecting to electricity and to the network, configuring routers and end-devices, setting passwords and defaults etc.). 
In fact, every \say{Do It Yourself} product with more than one sequential step, such as filling out government forms %, assembling artifacts 
or assembling complex items, complies with this notion.

Our goal is to make the service more efficient in several aspects: We want to be able to \textbf{attend more clients} with fewer human operators; we want to \textbf{minimize the time a client waits}, from the time that a help request is presented until the problem is solved; and we wish to \textbf{maintain the clients' satisfaction or even improve it}.

%Therefore, the challenge we address is how to build effective virtual agents and how to combine their deployment with human operators in order to achieve an efficient system, considering the two aforementioned aspects. Is it possible to obtain these two aspects together, or does the two-step process (a virtual agent that conveys the problem to a human operator) impel longer service times? 

\section{Model Presentation}
\label{sec:our_model}
In order to achieve the goals, we introduce \textbf{virtual agents} into the process. Our virtual agents are software entities that accept textual questions from the clients and provide them with advice to solve the problem. Of course, virtual agents cannot solve all of the problems that might arise, hence questions that agents fail to answer %satisfactorily 
are routed to a human operator.

We assume a service center with \textbf{a single human operator} who provides service to clients who perform sequential tasks. The clients contact the service center via \textbf{a chat channel}. When a help request arrives, it is routed to a virtual agent. There are enough virtual agents to attend calls from all of the clients simultaneously, without queuing.

When asked a question, a virtual agent checks whether it understands correctly: It returns %its own
a rephrased version of the question to the client and asks if this was what they meant. If the client responds affirmatively, the agent provides its answer to the question, then asks whether the answer satisfies the client. Another affirmative answer will end the session successfully. Alternately, if the client answers either question negatively, the agent routes the question to the queue of questions awaiting the human operator's attendance. 

The operator attends the questions waiting in the queue sequentially, one question at a time. When finishing with a question, he/she addresses the next one, if such exists. The human operator attends only to questions that the virtual agents failed to answer. 

A few notes on said process: First, we assume that the virtual agent is capable of identifying questions that it cannot answer. % and then route them %quickly  to the operator's queue. 
Thus it does not waste a lot of time contemplating questions beyond its capabilities. Second, in order to save time and to calm impatient clients, arriving questions are sent to the operator's queue in the order of their arrival to the system. While they are attended to by the virtual agents, they also advance in the queue. If, after trying, the virtual agent failed to answer a question, the client's place in the queue is as if it had arrived directly to the human operator's queue. Thus the time of the virtual agent's attendance is not \say{wasted time} for the clients, since they would have been waiting this time anyway: Instead of listening to \say{elevator music} while waiting, they chatted with the virtual agent in an attempt to get a quick answer, and even though this attempt did not pan out, they did not, in effect, lose time.  

\section{Theoretical Analysis}
\label{sec:Model_Analysis}
We divide the questions that clients ask into two types:
\begin{itemize}
    \item $\alpha$-questions: questions that the virtual agent knows how to answer. The fraction of $\alpha$-questions (out of the total number of questions) is $\alpha$. 
    \item $\beta$-questions: questions that the virtual agent cannot answer. The fraction of hard questions %out of the total number of questions 
    is $\beta = 1-\alpha$. 
\end{itemize}
It is assumed that the arrival rate of the questions is Poisson, and that the $\alpha$-questions are iid among them.

We call the environment with the virtual agent the \emph{hybrid} environment, and that with the human operator alone the \emph{pure human} environment.  

We now show that the hybrid system can service more clients while simultaneously also reducing the expected waiting times of the clients.

Let $c$ be the ratio of the number of clients in the hybrid environment to that of the pure-human environment. 
Let $s,s_{\beta}$, be the average (human) service time for questions in general, and for $\beta$ questions, respectively, and let $\epsilon$ be the time it takes the virtual agent to determine that the question is a $\beta$-question. Employing queuing theory we obtain:
\begin{theorem}
\label{th:queue}
Assuming $\epsilon \ll s_{\beta}$, so long as $c \leq {s\over {\beta s_{\beta}}}$, the expected waiting time in the hybrid system is less than that in the pure human system.
\end{theorem}
In the supplementary material we show that necessarily ${s\over {\beta s_{\beta}}}>1$.  So, for $1<c\leq {s\over {\beta s_{\beta}}}$ the hybrid environment allows for a greater number of clients, while simultaneously also reducing the waiting time.  
The proof of the theorem is provided in the supplementary material as well.

Section \ref{sec:Simulation} further explores the settings of the various parameters in order to achieve this win-win situation in a hybrid environment. Specifically, Figure \ref{fig:conditions} provides a comparison between different settings and exhibits the substantial gains that can be obtained.

%OLD: 
%Hence, with a cautious choice of system parameters, we can deploy our system using virtual agents such that \textbf{the number of clients that can be attended by a single human operator is higher} (Equation \ref{eq:n_with_Agents} and its derivatives) and still \textbf{the expected waiting time in queue for clients is shorter} (Equation \ref{eq:waitingTimes}) and \textbf{the operator's utilization ratio is lower} (Equation \ref{eq:rho}) than in a scenario without virtual agents. This is an interesting situation, and in the following sections we explore the practical opportunities it suggests.  

%\end{enumerate}

\section{Designing and Integrating the Virtual Agents: Methodology and Human Evaluation}
\label{sec:Designing and Integrating the Virtual Agents: Methodology and Human Evaluation}
In order to test the model and its applicability, we implemented an environment that enables the performance of human experiments. In the following sections we describe the environment, the agents and the experiments we performed.

\subsection{Environment}
\label{sec:Environment}
As a basis, we chose the computer game \say{PC Building Simulator}, a simulation-strategy video game produced by The Irregular Corporation and Romanian independent developer Claudiu Kiss (reviewed in \cite{PCBuildingSimulator}). The game is centered around owning and running a workshop which builds and maintains PCs. %It is available via the \say{Steam} video game digital distribution service \cite{Steam}.
In the original game, the players -- our clients -- are supposed to perform diverse tasks in order to build and fix PCs: They need to communicate with service seekers (accept PCs for service, ask for preferences, collect fees), to technically maintain PCs (disassemble, change parts, clean motherboards, install new gear, reassemble), to install anti-virus and other software products, to order parts from suppliers and pay for them, etc. 

%\begin{figure}[ht]
%\centering
%\frame{\includegraphics[width=0.47\textwidth]{PCSimulator.PNG}}
%\caption{\label{fig:PCSimulator} A screenshot of the PC Building Simulator game. The open case of the PC in repair progress is in the center. Tasks to perform are listed in red on the right.}
%\end{figure}

To the original game (which instantiates the tasks the client has to perform), we added a Navigator -- an interactive service that accompanies the player, shows the tasks' instructions (i.e., the user manual), and asks the players to report if they successfully performed a step or if help is needed. In the latter case, they can pose questions to an experienced human operator. % (Fig. \ref{fig:guidme}). 
The questions and answers are communicated via a textual chat channel. The human operator can also remotely watch the client's screen in order to better understand the situation and the problem. In the experimental scenario we used there were 107 steps, i.e. 107 different tasks that the Navigator guides the player to perform, in a specific predefined order.

%\begin{figure}[ht]
%\centering
%\frame{\includegraphics[width=0.37\textwidth]{guideme.PNG}}
%\caption{\label{fig:guidme} The Navigator -- the client reports on successful completion of tasks, and alternately ask for the operator's help.}
%\end{figure}

%\color{red}The aforementioned environment was used to exhibit a methodology for the implementation of two building blocks of the system: a virtual agent capable of answering (some) questions in place of a human operator; and an intelligent operator interface, designed to shorten the time needed by a human operator to answer hard questions that the virtual agent fails to answer. In terms of the model presented in Section \ref{sec:Model_Analysis}, the first block's goal is to introduce the virtual agents and to increase $\alpha$ -- the fraction of questions answered by the agent, while the second block's goal is to decrease $s_\beta$ -- the expected service time by the human. These two blocks, when combined together, may facilitate the circumstances to implement efficient real-world systems that realize the suggested model. \color{black}

%  the   and conduct two different experiments, with two separate goals. In the first, described in Section \ref{sec:Building_a_Virtual_Agent}, we developed a methodology to implement the virtual agent to automatically answer the clients' questions, and tested its abilities.  These agents would later help clients instead of the human operator, as we suggested in our model. The second experiment, in Section \ref{sec:Building_an_Improved_Operator's_Interface}, is 

\subsection{The Virtual Agent}
\label{sec:Building_a_Virtual_Agent}
The aforementioned environment was used to implement a hybrid service system in which we perform experiments with human subjects. We based our agent solely on the transcripts of the communication between clients, the Navigator (described in Section \ref{sec:Environment}) and the human operator.  The agent is not provided with any additional domain knowledge. When a client asks a question using the Navigator, a virtual agent checks whether it has a similar question (and a corresponding answer) in its questions-and-answers DB. If the agent finds a similar question, it asks the client whether this is  what he/she meant. If so, then the agent responds with the corresponding answer from the DB. Otherwise, no further attempts are made by the agent to answer the question (in order to keep  $\epsilon$ small), and the agent passes the question to the human operator, who answers it. Naturally, the (new) human answer (together with the question) is added to the DB, and will be used by the virtual agent in the future. This way the DB is built and developed incrementally, based on human operators’ answers. In order to avoid the cold start problem, we used the Bag of Words method (Goldberg 2017) for the first steps.

\begin{figure}[t]
\centering
\includegraphics[width=0.47\textwidth]{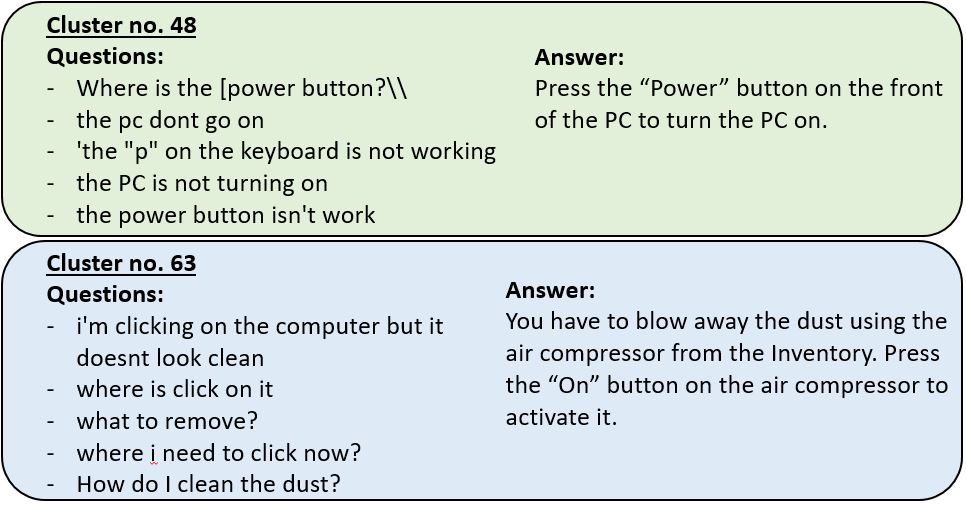}
\caption{\label{fig:clusters} Two example clusters, built from the chat communication between subjects (playing the role of clients) and an experimenter (in the role of an operator) in the experiment. Note that questions appear in their raw state, just the way they were written by clients.}  
\end{figure}

It should be noted that, in this process, for each question asked by a client and for each answer provided by the system, the clients indicate whether this was their intention and whether the provided response indeed answered their question. Through this process, in addition to answering the clients' questions, we get a set of validated clusters of questions (i.e., each cluster contains various phrasings to a single question) and a validated answer to each cluster. Figure \ref{fig:clusters} shows two examples of such clusters. %, presenting the list of questions and the operator's answers. 

\subsection{Human Evaluation}

{\bf Virtual Agent Deployment Experiment. }
The goal of the first experiment was to explore the process of building the virtual agent and its ability to answer customers' questions. We recruited 41 subjects who played the role of clients (19 males, 22 females, average age 26.2 years old), and an experimenter played the role of an operator. The sessions were performed via the Internet, similar to a real-world situation in which clients are in their homes or offices and request help from a remote service center. 

\begin{figure}[t]
\centering  
\includegraphics[width=0.47\textwidth]{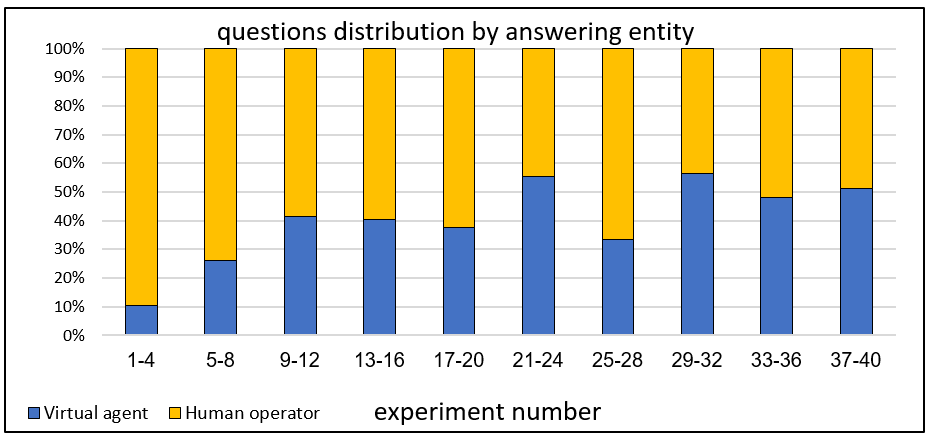}
\caption{\label{fig:experimentsBarChart} Distribution of questions according to answering entity -- human operator vs. virtual agent -- first 40 sessions.}
\end{figure}

Figure \ref{fig:experimentsBarChart} presents the distribution of questions according to the answering entity -- the human operator or the virtual agent. In the first sessions, the virtual agent had a poor answering ability (since the DB is sparse), therefore most of the questions were passed on to, and answered by, the human operator. As data accumulated, the agent developed a growing ability to answer questions, and by the 30th session $\alpha$ was already around $0.5$, i.e. agents could satisfactorily answer about half of the clients' questions. After performing 41 sessions and accumulating data, we applied several pre-trained LM models in order to explore their ability to identify the right cluster for a given question. Using the paraphrase-MiniLM-L6-v2 model \cite{reimers-2019-sentence-bert} with the raw clusters we achieved $81\%$ accuracy and $74\%$ precision. It should be noted that, in calibrating the model's behaviour, we maintained a cautious policy in order not to annoy the clients: The agent is allowed to have only one attempt to answer the client's question, and if this attempt fails the agent refrains from  trying again. Furthermore, when considering the agent's suggestion, we rather misdetect questions (i.e., pass them on to the human operator when not positively affirmative that we correctly identified the client's intention) than falsely alarm the client with a wrong suggestion. This policy is performed by setting a relatively high threshold that must be met when considering whether to allow the agent to offer its suggestion to the client, or to pass the question directly to the human operator.

{\bf User Satisfaction Study. }
The first experiment's results show that it is feasible to build an effective virtual agent in a short amount of time using real-world existing processes (such as an existing pure human service center). Nevertheless, the quality of service, and specifically the clients' experience of the hybrid configuration, could not be measured in this design of experiment. To this end, we designed a second experiment, which aimed to compare clients' satisfaction in two conditions: (1) Pure human  service, in which a single human operator attends to 11 clients, and (2) Hybrid service, in which a single human operator, assisted by virtual agents, attends to 18 clients. In both conditions, one of the clients was played by a human recruit and the other clients were virtual, i.e., they were place holders that waited in the queue and took the human operator's attention in order to make the experience of the real human client authentic. The amount of time each virtual client needed to be "served" was generated using the real-world data we collected in the first experiment about the needs of the clients and the time it takes the operator to serve them. Similarly, the values of 11/18 clients for the different conditions were chosen according to our experience in the first experiment, in order to produce similar waiting times in the queue for the human operator: Obviously, in the second condition some of the clients' questions are answered by the virtual agents, thus easing the load on the human operator. We wanted the clients in the experiment to experience equal waiting times for the human operator in both conditions, in order to measure the clients' satisfaction not as a result of shorter waiting times in Condition 1, but in relatively equal situations. 20 recruits took part in the experiment as clients (12 males, 8 females, average age 21.4 years old). At the end of each session, the human clients were asked to grade their satisfaction from the service they received in three categories: The waiting times, the quality of service in general, and the quality of information they received from the help desk. 

\begin{figure}[ht]
\centering
\includegraphics[width=0.47\textwidth]{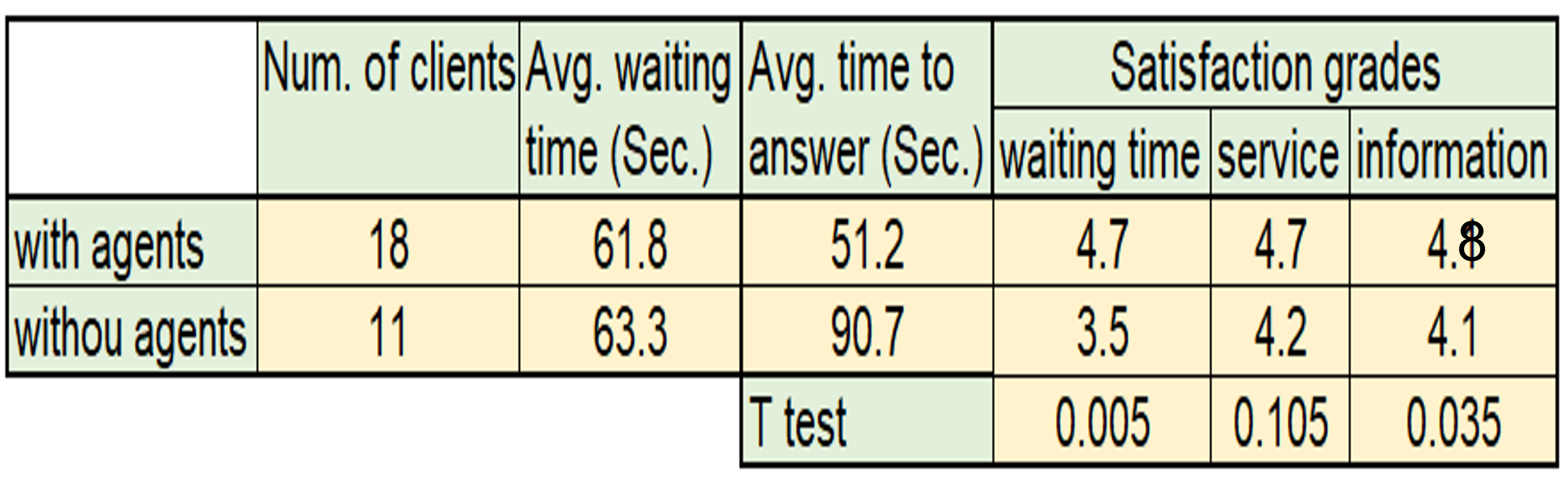}
\caption{\label{fig:SecondExperiment} Results of the user satisfaction study.}
\end{figure}

The results of the second experiment are presented in Figure \ref{fig:SecondExperiment}. It can be seen that although the number of simultaneous clients was higher in Condition 1 (18 "with agent" vs. 11 "without agent"), the average waiting time for a human operator was similar in both conditions (61.8 vs 63.3), and the quality of service was higher in Condition 1: Clients received answers to their questions significantly faster (51.2 vs 90.7), finished the game-steps sooner and reported higher satisfaction grades. 

Demographics and other data regarding the subjects in both experiments can be found in the supplementary material. The study protocol was approved by the Institutional Review Board (IRB) of the anonymous university. 

Combining the results of both experiments, we conclude that it is possible to train an agent to answer a significant number of clients' questions, and to use this agent to implement the hybrid model that enables an increase in the number of clients while maintaining, and even improving, the quality of service.   

The usability of this kind of agent and the quality of its answering capacity are further contemplated in the simulations presented in Section \ref{sec:Simulation}.

\section{Simulation of Virtual Service Center}
\label{sec:Simulation}

In Section \ref{sec:Model_Analysis} we presented a theoretical analysis, showing that it is possible to achieve a win-win situation using a hybrid environment. Nevertheless, as we mentioned in the closing paragraph of that section, this analysis does not explore the practicality of the model, i.e., whether it can work with realistic parameters and provide proper service to actual clients.

In order to test the model and the relations between parameters, we built a simulator that implements the system as described in Section \ref{sec:our_model}. The input parameters are $n$ and $\alpha$, and the distributions of $s_\alpha$, $s_\beta$, $\bar{\lambda}$ and $\epsilon$. We studied the behaviour of the system with different values of parameters, and tried to understand the possible improvements in performance using the tools mentioned in Section \ref{sec:Building_a_Virtual_Agent}. Note that in the simulations we also explored the effect of shortening the response time of human operators, an issue that was investigated but not elaborated on in this paper due to space restrictions.

\begin{figure*}[ht]
\centering
\includegraphics[width=1\textwidth]{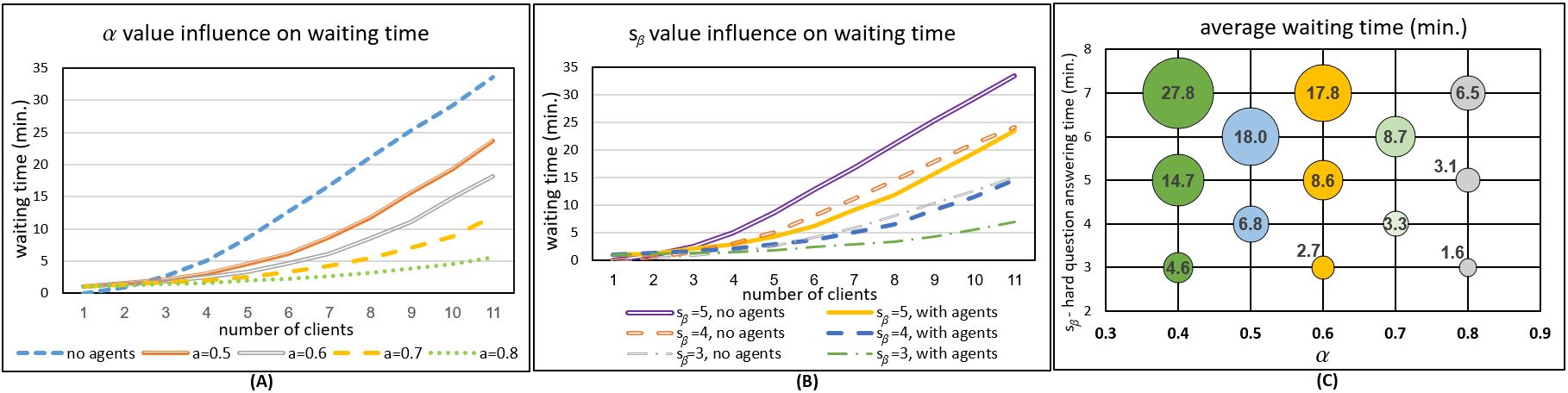}
\caption{\label{fig:combined} (A) Expected waiting time for a client with a hard question, comparing different values of $\alpha$. In all cases, $s_\beta \sim \mathcal{N}(5,1)$, $s_\alpha \sim \mathcal{N}(1,0.2)$, $\epsilon \sim \mathcal{N}(0.5,0.1)$, $\bar{\lambda}\sim $ Pois$(0.1)$.  (B) Expected waiting time for a client with a hard question, comparing different values of $s_\beta$. In all cases, $\alpha=0.5$, $s_\alpha \sim \mathcal{N}(1,0.2)$, $\epsilon \sim \mathcal{N}(0.5,0.1)$, $\bar{\lambda}\sim $ Pois$(0.1)$.  (C) Expected waiting time (minutes) for a client with a hard question for combinations of $\alpha$ and $s_\beta$, in a hybrid environment. In all cases, $n=8$, $s_\alpha \sim \mathcal{N}(1,0.2)$, $\epsilon \sim \mathcal{N}(0.5,0.1)$, $\bar{\lambda}\sim $ Pois$(0.1)$.}
\end{figure*}

\begin{figure}[t]
\centering
\includegraphics[width=0.47\textwidth]{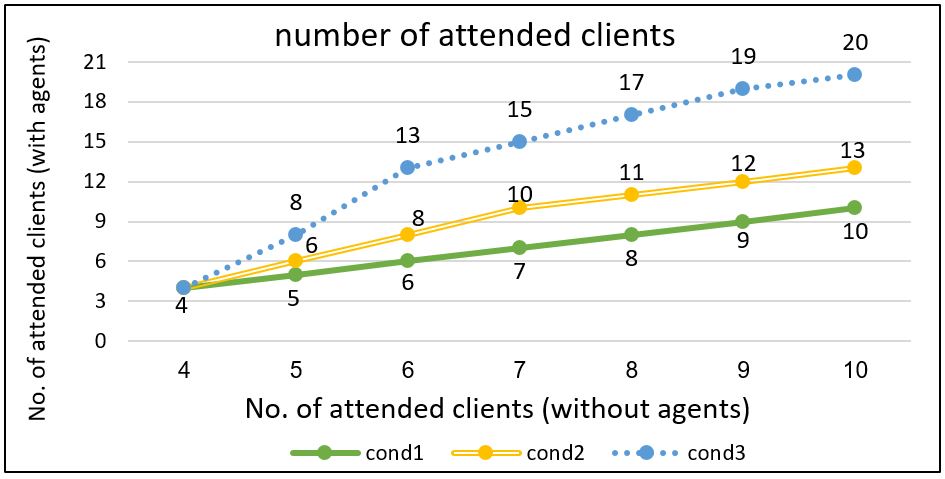}
\caption{\label{fig:conditions} The number of clients that a human operator can attend in a hybrid environment, relative to the number of clients that a human operator attends in a pure human environment without increasing the waiting time. In all three conditions, $s_\alpha \sim \mathcal{N}(1,0.2)$, $\epsilon \sim \mathcal{N}(0.5,0.1)$, $\bar{\lambda}\sim $ Pois$(0.1)$. Cond1: $\alpha=0.2$, $s_\beta \sim \mathcal{N}(3.5,1)$. Cond2: $\alpha=0.5$, $s_\beta \sim \mathcal{N}(3.15,1)$. Cond3: $\alpha=0.7$, $s_\beta \sim \mathcal{N}(2.8,1)$.} 
\end{figure}

Figure \ref{fig:combined} presents results of multiple simulations. In order to make the simulations realistic, we extracted actual parameters from the experiments we conducted. For example, we empirically learned that the average rate of questions that a client asks is 0.1 questions per minute, hence we set $\bar{\lambda}$ to have a Poisson distribution with parameter $0.1$ (that is, $\bar{\lambda}\sim $ Pois(0.1)). We also learned that $\alpha=0.5$ is a realistic value after a short training process (see Figure \ref{fig:experimentsBarChart}), and that $3.5$ minutes are a reasonable start value for $s_\beta$ (Figure \ref{fig:conditions}). 

Figure \ref{fig:combined}A demonstrates the decrease in the expected waiting time for a client with a hard question by improving the agents' ability to supply answers to questions. Assuming that an expected waiting time of longer than 5 minutes is not acceptable as a decent service, we see that without virtual agents a human operator can attend no more than 4 clients simultaneously. With agents that can answer half of the questions ($\alpha=0.5$), the operator can attend 6 clients. Increasing $\alpha$ to $0.8$ enables a single human operator to attend 11 clients simultaneously.

%\begin{figure}[ht]
%\centering
%\includegraphics[width=0.47\textwidth]{figwaitingtimeaccordingtoalpha.JPG}
%\caption{\label{fig:alpha} Average waiting time for a client with a hard question, comparing different values of $\alpha$. In all cases, $s_\beta \sim \mathcal{N}(5,1)$, $s_\alpha \sim \mathcal{N}(1,0.2)$, $\epsilon \sim \mathcal{N}(0.5,0.1)$, $\bar{\lambda}\sim $ Pois$(0.1)$.}
%\end{figure}

Fig. \ref{fig:combined}B demonstrates the decrease in the expected waiting time for a client with a hard question by reducing the time it takes a human operator to answer hard questions. A human operator without agents, who needs about 5 minutes to answer a hard question, can attend up to 4 clients at once. Assisted by virtual agents (even \say{weak} ones, with $\alpha=0.5$), the operator can attend 6 clients, and when improving the interface and decreasing the answer time of hard questions to 3 minutes, the operator may attend 10 clients simultaneously. In additional simulations we found that if we combine both processes -- increasing $\alpha$ to $0.8$ \emph{and} decreasing $s_\beta$ to $3$ minutes -- a single operator can attend 20 clients in a hybrid environment and still maintain an expected waiting time of lower than $5$ minutes ($4.96$ min.), whereas only $6$ clients can be attended to at once in a pure human environment.  

%\begin{figure}[ht]
%\centering
%\includegraphics[width=0.47\textwidth]{figwaitingtimeaccordingtosh.JPG}
%\caption{\label{fig:sh} Average waiting time for a client with a hard question, comparing different values of $s_\beta$. In all cases, $\alpha=0.5$, $s_\alpha \sim \mathcal{N}(1,0.2)$, $\epsilon \sim \mathcal{N}(0.5,0.1)$, $\bar{\lambda}\sim $ Pois$(0.1)$.}
%\end{figure}

Fig. \ref{fig:combined}C demonstrates expected waiting times for a hard question for several combinations of $\alpha$ and $s_\beta$ values, when attending $8$ clients simultaneously in a hybrid environment. The sizes of the bubbles represent the expected waiting times, and it shows that waiting times may be significantly reduced simply by changing these parameters.

%\begin{figure}[ht]
%\centering
%\includegraphics[width=0.47\textwidth]{figpoints.jpg}
%\caption{\label{fig:points} Expected waiting time (minutes) for a client with a hard question for combinations of $\alpha$ and $s_\beta$. In all cases, $n=8$, $s_\alpha \sim \mathcal{N}(1,0.2)$, $\epsilon \sim \mathcal{N}(0.5,0.1)$, $\bar{\lambda}\sim $ Pois$(0.1)$.}
%\end{figure}

In Section \ref{sec:Model_Analysis} we claimed that it is possible to adjust the hybrid environment's parameters to provide a win-win situation, where a single operator can serve more clients and at the same time the clients experience shorter waiting times. Figure \ref{fig:conditions} presents simulations of 3 conditions and compares the number of clients that can be attended in a pure human environment to the number of clients that can be attended in a hybrid environment. For every value of $n$ (no. of clients) in a pure human environment on the x-axis, we matched the number of clients that can be attended in a hybrid environment with a lower expected waiting time per client. In Condition 1 ($\alpha=0.2$, $s_\beta \sim \mathcal{N}(3.5,1)$), the hybrid environment can attend the same number of clients as the pure human one ($c=1$). In Condition 2, with improved answering capability of the agents ($\alpha=0.5$) and a shorter time to answer hard questions ($s_\beta \sim \mathcal{N}(3.15,1)$), the hybrid environment is capable of attending slightly more clients than the pure human one ($c \approx1.3$). In Condition 3, with these parameters further improved ($\alpha=0.7$, $s_\beta \sim \mathcal{N}(2.8,1)$), the hybrid environment supports a significantly higher number of clients per human operator ($c \approx 2$). In all conditions, other parameters are fixed: $s_\alpha \sim \mathcal{N}(1,0.2)$, $\epsilon \sim \mathcal{N}(0.5,0.1)$, $\bar{\lambda}\sim $ Pois$(0.1)$.     

%In the analysis of Section \ref{sec:Model_Analysis} and in the aforementioned simulations we assumed a single human operator who attends $n$ clients. It definitely simplifies the analysis, but having a team of $k$ operators who attend $nk$ clients may allow them to cope even better with extreme cases and decrease the expected waiting time. Results of simulations that explore this issue are presented in the supplementary material. 
\subsection{Teams of Human Operators} 
\begin{figure}[h]
\centering
\includegraphics[width=0.47\textwidth]{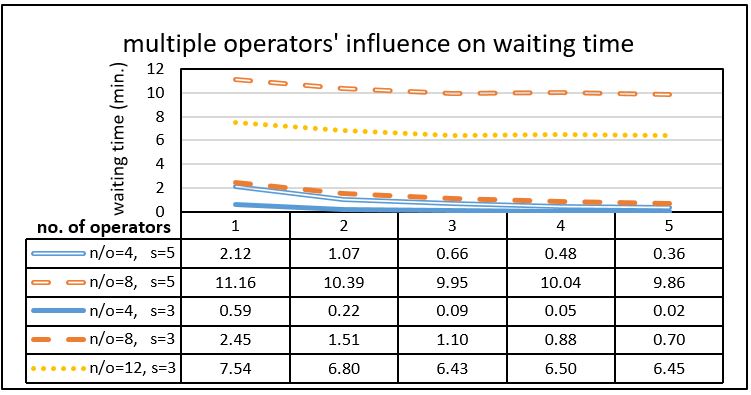}
\caption{\label{fig:multipleOperators} The influence of the number of operators in the team on the expected waiting time (in minutes). $n/o$ is the number of clients per a single human operator. In all cases, $\alpha=0.5$, $s_\alpha \sim \mathcal{N}(1,0.2)$, $\epsilon \sim \mathcal{N}(0.5,0.1)$, $\bar{\lambda}\sim $ Pois$(0.1)$.} 
\end{figure}

In the analysis of Section \ref{sec:Model_Analysis} and in the aforementioned simulations we assumed a single human operator who attends $n$ clients. It definitely simplifies the analysis, but having a team of $k$ operators who attend $nk$ clients may allow them to cope better with extreme cases and decrease the expected waiting time. Results of simulations that explore this issue are presented in Figure \ref{fig:multipleOperators}, which shows the expected waiting time for a client for various sizes of operator teams. Every line represents a fixed clients-to-operators ratio ($n/o$), e.g. $n/o=4$ is checked in the cases of 1 operator with 4 clients, 2 operators with 8 clients, 3 operators with 12 clients and so on. 

A slight decrease in the average waiting time is visible mostly in the change from a single operator to two operators, because momentary load can be divided between the operators.  Further increasing the number of operators has only negligible influence. It can also be seen that the decrease in the expected waiting time is quite steady (between $0.57$ to $1.76$ for all cases), and is not proportional to the value of the waiting time. Thus it is significant in cases where the expected waiting time is low (e.g. $n/o=4, s_\beta=5$), but almost negligible when the expected waiting time is high (e.g. $n/o=8, s_\beta=3$). 
These findings led us to concentrate our analysis on the single operator scenario, understanding that the results can be extrapolated for higher values.

\section{Conclusions and Future work}
In this work we presented, analysed and simulated a hybrid environment combining human operators with virtual agents to provide customer service for complex tasks. We used queue theory, experiments and simulation to prove and demonstrate that, using NLP techniques, it is possible to implement this method in a way that yields a shorter client waiting time while serving more clients than is possible in a pure human environment.

In our experiments we focused on question-answering methods. In this context, it is possible to deploy other methods, such as conversational virtual agents \cite{motger2021conversational,montenegro2019survey}. Nevertheless, conversational agents may increase the time it takes an agent to notice that it can't provide an answer (i.e., $\epsilon$ in our model), thus increasing the total waiting time. We examined the relations between $\epsilon$ and $\alpha$ regarding expected waiting time but could not expand on this issue here due to lack of space.

Another issue that should be explored is the optimal balance between an agent's effort to answer a question by itself and transferring it to a human operator. In our experiment, the agent performed one attempt to provide an answer to the client, and if this attempt failed it passed the question on to the operator. There are other approaches that may be considered. For example, the agent may perform up to a constant $n>1$ attempts, or attempt to offer all the questions in the DB that were graded above the threshold. These schemes may increase the value of $\alpha$, but they might also annoy the clients and decrease their satisfaction. This issue may be explored and presented in future works.

\section*{Acknowledgement}
This research was funded in part by JPMorgan Chase \& Co. Any views or opinions expressed herein are solely those of the authors listed, and may differ from the views and opinions expressed by JPMorgan Chase \& Co. or its affiliates. This material is not a product of the Research Department of J.P. 
Morgan Securities LLC. This material should not be construed as an individual recommendation for any particular client and is not intended as a recommendation of particular securities, financial instruments or strategies for a particular client. This material does not constitute a solicitation or offer in any jurisdiction.

%\clearpage

\bibliographystyle{named}
%\bibliography{handoff}

\section{Appendix}

\section{Proof of Theorem 1}

\paragraph{Parameters and Notation.}
\begin{itemize}
    \item Number of clients in the system:
    \begin{itemize}
        \item $n$: number of clients in the pure human environment. 
        \item $\hat{n}=c\cdot n$: number of clients in the hybrid environment.
        \end{itemize}
    
    \item Arrival rates: It is assumed that arrivals follow a Poisson distribution, with the following rates:  
    \begin{itemize}
        \item $\bar{\lambda}$: arrival rate of questions per client.
        \item $\lambda=n\bar{\lambda}$: overall arrival rate of questions in the human environment. 
        \item $\hat{\lambda}=c\beta\lambda$: arrival rate of $\beta$-questions in the hybrid environment. 
    \end{itemize}
    
    \item Service times (all service times are per question)
    \begin{itemize}
        \item $s$, $\sigma_s$: average and standard deviation of the service time by human in the pure human environment. 
        \item $s_{\beta}, \sigma_{s_{\beta}}$: average and standard deviation service time by human for $\beta$ questions. We denote $\tau = \frac{s_{\beta}}{s}$.
        \item $s_{\alpha}$, $s_{\alpha}^A$: average service time by human and agent, respectively, for $\alpha$-questions.
        We assume $s_{\alpha}^A\approx s_{\alpha}$.
% the human service time.
    \end{itemize}
    \item  Utilization: The fraction of time the queue is non-empty.
    \begin{itemize}
        \item $\rho=\lambda s$: utilization in the pure human environment. 
        \item $\hat{\rho}=\hat{\lambda}s_{\beta}$: utilization in the hybrid environment. 
    \end{itemize}
    
    %\item Expected time in queue (not including service time)
    %\begin{align}
     %   W_q = \frac{\lambda \sigma_s^2}{2(1-\rho)}+\frac{\rho^2}{2\lambda(1-\rho)}
    %\end{align}
\end{itemize}

\begin{itemize}
    
    \item $\epsilon$: The time it takes for a virtual agent to notice that it cannot answer a question.     

\end{itemize}

\paragraph{Relationships.} Note that 
\begin{align}
    s = \beta s_{\beta}+(1-\beta)s_{\alpha}.
    \label{eq:s}
\end{align}
\begin{align}
\mbox{so, } s> \beta \tau s & \mbox{ and }    \beta \tau < 1.  \label{eq:betatau}
\mbox{ Also note that }
    \sigma_{s_{\beta}} < \sigma_s, 
    \end{align}
since we may assume that the service times of the $\beta$-questions are more similar to each other than to the $\alpha$-questions. 

\paragraph{Waiting Times.}
\label{sec:analysis - waiting times}
We shall employ the M/G/1 queuing model. This model assumes a Poisson arrival rate (as do we), but the distribution of the service rate is arbitrary. 

When employing the M/G/1 model in the pure human environment, 
the expected waiting time in the queue (not including service time) is:
%\begin{itemize}
    %\item Without agent:
\begin{align}
\label{eq:w}
        W = \frac{\lambda \sigma_s^2}{2(1-\rho)}+\frac{\rho^2}{2\lambda(1-\rho)}
\end{align}
For the hybrid environment, $\alpha$-questions are answered immediately, so only $\beta$-questions wait in the queue. Hence, the expected waiting time per question in the hybrid system is
%\item With agent:
\begin{align}
        \hat{W} &= \beta\left( \frac{\hat{\lambda} (\sigma_{s_{\beta}})^2}{2(1-\hat{\rho})}+\frac{\hat{\rho}^2}{2\hat{\lambda}(1-\hat{\rho})}+\epsilon\right)
        \label{eq:whata}
\end{align}
The first two addends in the parentheses are the expected waiting time of $\beta$ questions, and the last addend is the time it takes to notice that the question is a $\beta$-question. Thus the sum in the parentheses bounds the expected waiting time of a $\beta$-question.\footnote{Here we calculated as if the entire time it takes the agent to notice that it cannot answer the question is wasted. In practice, as detailed in Section 3, if the queue is not empty, the question is placed in the queue immediately upon arrival, and the virtual agent attends to it while it waits in the queue.  So, in such cases, which happen almost a $\hat{\rho}$ fraction of the time, the extra $\epsilon$ need not be added.} 
However, $\beta$-questions are only a $\beta$-fraction of the questions, the rest having zero waiting time. Therefore, the overall expected waiting time is $\beta$ times this sum.

Rearranging \eqref{eq:whata}, we have:
\begin{align}
         \hat{W}  &=  \frac{c\beta^2\lambda (\sigma_{s_{\beta}})^2}{2(1-c\beta\tau\rho)}+\frac{(c\beta\tau)^2\rho^2}{2c\lambda(1-c\beta\tau\rho)}+\beta\epsilon  \label{eq:wa}
\end{align}

%\end{itemize}

%We want: $\hat{W}_q \leq W_q$.

%\paragraph{Simplifications:}
%\begin{enumerate}
\noindent If $\beta\epsilon$ is small, then:
\begin{align}
\label{eq:wa-sim1}
        \hat{W}  &\approx \frac{c\beta^2\lambda (\sigma_{s_{\beta}})^2}{2(1-c\beta\tau\rho)}+\frac{(c\beta\tau)^2\rho^2}{2c\lambda(1-c\beta\tau\rho)}
\end{align}
Note that \eqref{eq:wa-sim1} in monotonically increasing in $c\beta\tau$.
By assumption (of Theorem 1), ${\hat{n} \over n}\leq {s\over {\beta s_{\beta}}}$, so $c\beta\tau\leq 1$. In the right-hand side of \eqref{eq:wa-sim1} we may therefore replace $c\beta\tau$ by $1$ and only increase the value. We thus obtain:
\begin{align}
\label{eq:wa-sim2}
        \hat{W}  &\leq  \frac{c\beta^2\lambda (\sigma_{s_{\beta}})^2}{2(1-\rho)}+\frac{\rho^2}{2c\lambda(1-\rho)}.
\end{align}
We have that 
\eqref{eq:wa-sim2}$<$\eqref{eq:w}, since:
\begin{align*}
    &c\beta^2<c\beta\tau =1  \ \ \ \mbox{and} \ \ \
    \sigma_{s_{\beta}}<\sigma_s
\end{align*}
Therefore  
\begin{equation}
\label{eq:waitingTimes}
\hat{W} < W. 
\end{equation}

%\subsection{System Capacity}
%We now show that the hybrid system allows an increase in the number of clients, while maintaining the same utilization. 
%\begin{align}
%    \rho =&\hat{\rho} \ \ \  \Rightarrow  \ \ \  %\nonumber 
%    s\lambda = c\beta \lambda s_{\beta} = c\beta \lambda s %\frac{s_{\beta}}{s} \ \  \Rightarrow \ \ \
%    1 \ =  \  c\beta \frac{s_{\beta}}{s} \  \Rightarrow \nonumber \\
%    c = &\frac{s}{\beta s_{\beta}} = \frac{\beta s_{\beta} + %(1-\beta)s_{\alpha}}{\beta s_{\beta}} \ 
%    =  \ 1 + \frac{1-\beta}{\beta}\cdot \frac{s_{\alpha}}{s_{\beta}}
%\label{eq:cone} 
%\end{align}
%From Equation \ref{eq:cone} we see that one can increase the number of clients ($c > 1$) while maintaining the same utilization. Furthermore, $c$ increases as $\beta$ decreases (i.e., as the ratio of questions that the agents can answer is higher).
%\subsection{Waiting Times}

%---------------------------------------------------------

\section{Experiment 1 - Subjects' Data}
\begin{figure}[h]
\centering
\includegraphics[width=0.25\textwidth]{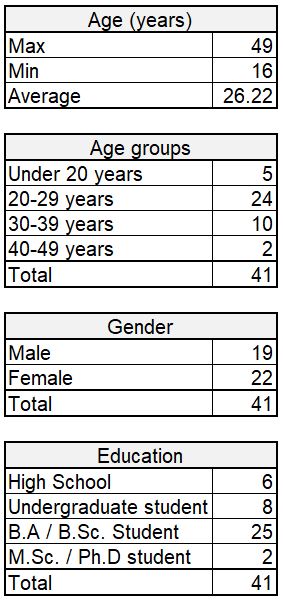}
\caption{\label{fig:exp 1 demographics} Demographic data of the human subjects who took part in Experiment 1.} 
\end{figure}

In experiment 1 we recruited subject from the general population via the internet. Recruits were paid for the participation. 41 participants took part in the experiment.
Figure \ref{fig:exp 1 demographics} presents the demographic data of the participants.

\hfill \break
\hfill \break
\hfill \break
\hfill \break
\hfill \break
\hfill \break
\thispagestyle{empty}

\newpage
\section{Experiment 2 - Subjects' Data}
\begin{figure}[h]
\centering
\includegraphics[width=0.28\textwidth]{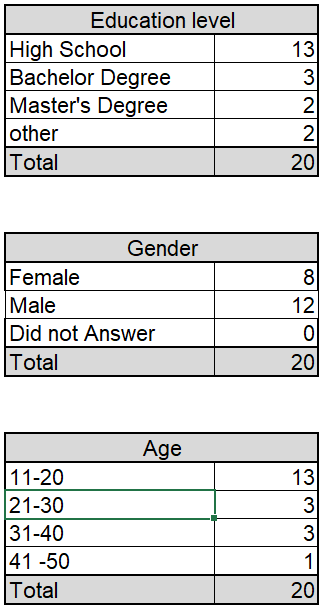}
\caption{\label{fig:exp 2 demographics} Demographic data of the human subjects who took part in Experiment 2.} 
\end{figure}

In experiment 2 we recruited subject from the general population via the internet. Recruits were paid for the participation. 20 participants took part in the experiment. Figure \ref{fig:exp 2 demographics} presents the demographic data of the participants.

\end{document}